\documentclass[12pt]{article}
\usepackage[utf8]{inputenc}
\usepackage{url}
\usepackage{geometry}
\usepackage{graphicx}
\usepackage{multirow}
\usepackage{caption}
\usepackage{subcaption}

\usepackage{enumitem}
\newcommand{\subscript}[2]{$#1 _ #2$}
\usepackage{arydshln}

\title{Speed estimation evaluation on the KITTI benchmark based on motion and monocular depth information}
\author{R\'obert-Adrian Rill\textsuperscript{1,2}}
\date{\footnotesize%
\textsuperscript{1}Faculty of Informatics, E\"otv\"os Lor\'and University, Hungary\\%
\textsuperscript{2}Faculty of Mathematics and Computer Science, Babe\c{s}-Bolyai University, Romania
}

\begin{document}
\sloppy
\maketitle

\begin{abstract}
In this technical report we investigate speed estimation of the ego-vehicle on the KITTI benchmark using state-of-the-art deep neural network based optical flow and single-view depth prediction methods. Using a straightforward intuitive approach and approximating a single scale factor, we evaluate several application schemes of the deep networks and formulate meaningful conclusions such as: combining depth information with optical flow improves speed estimation accuracy as opposed to using optical flow alone; the quality of the deep neural network methods influences speed estimation performance; using the depth and optical flow results from smaller crops of wide images degrades performance. With these observations in mind, we achieve a RMSE of less than 1 m/s for vehicle speed estimation using monocular images as input from recordings of the KITTI benchmark. Limitations and possible future directions are discussed as well.
\end{abstract}

\section{Introduction}
The significant progress in recent years makes deep learning solutions attractive in automotive industry applications~\cite{yao2018machine}. In intelligent transportation systems, self-driving cars, advanced driver-assistance systems (ADAS) vehicle speed is one of the most important parameters for vehicle control and safety considerations. Velocity can be measured using wheel speed sensor, Inertial Navigation System (INS) or Global Positioning System (GPS). Although these methods can achieve high accuracy, they also have limitations. Speed sensors involve a trade-off between accuracy and cost. INS suffers from integration drift, i.e. small errors accumulate over time. GPS is prone to signal interference and loss in blocked areas, and may provide unreliable data when accelerating or decelerating. Some researchers also propose model-based approaches (see, e.g.,~\cite{qimin2017acost-effective}), which, however, may be significantly affected by incorrect parameter estimates.

Other methods for automotive sensing technologies, including speed estimation, are Radio Detection and Ranging (RADAR) or Light Detection and Ranging (\mbox{LiDAR}) systems that use radio frequency or laser signals, respectively~\cite{abuella2018vildar}. Although both systems are popular in the automotive industry, they have deficiencies. LiDAR systems can identify fine details of the 3D environment, but are greatly affected by unfavorable weather conditions and more importantly they are costly to produce and maintain. In contrast, RADAR systems are more robust, lightweight and cheap, but have much lower accuracy and resolution.

To address the limitations of conventional speed estimation methods, computer vision based approaches have become attractive alternatives in recent years. In this work we present a simple monocular vision-based speed estimation approach that exploits the recent advances in deep learning-based optical flow and monocular depth prediction methods. Optical flow is the pattern of apparent motion of objects in a visual scene caused by the relative motion between an observer and the scene. Monocular depth estimation aims to obtain a representation of the spatial structure of a scene by determining the distance of objects from a single image. The two problems are fundamental in computer vision and represent highly correlated tasks (see, e.g.,~\cite{zou2018dfnet}). We rely on the intuition that the magnitude of optical flow is positively correlated with the moving speed of the observer and that objects closer to the camera appear to move faster than the more distant ones. We evaluate different schemes of our method on a representative subset of the KITTI dataset~\cite{Geiger2012CVPR, Geiger2013IJRR}, and achieve a RMSE of less than 1 m/s.

The rest of the paper is organized as follows. Section~\ref{sect:related} presents related works providing a background and motivation for our study. Section~\ref{sect:methods} describes the KITTI dataset and the deep learning methods used in the present work, and details our speed estimation pipeline. The quantitative and qualitative results are presented in Section~\ref{sect:results}. Finally, Section~\ref{sect:discussion_conclusion} discusses the results and limitations, highlights future directions and concludes our work.

\section{Related work}\label{sect:related}
Vision based approaches represent a promising direction for vehicle speed estimation that may replace or complement traditional methods. Most of the works are concerned with estimating the velocity of the vehicles in traffic using a camera mounted for traffic surveillance. These methods involve different image processing techniques: background extraction~\cite{dogan2010real, temiz2012real, anilraoyg2015real-time, luvizon2016avideo-based}, image rectification~\cite{dogan2010real, temiz2012real, anilraoyg2015real-time, kumar2018asemi-automatic}, detecting and tracking reference points~\cite{dogan2010real, temiz2012real, luvizon2016avideo-based} or centroids~\cite{anilraoyg2015real-time} over successive frames, converting the velocity displacement vectors from the image to the real-world coordinate system. The state-of-the-art results of deep learning in vision tasks makes object detection and tracking also suitable in the task of speed estimation~\cite{kumar2018asemi-automatic}. Alternatively, locating license plates on the vehicles is another promising direction~\cite{luvizon2016avideo-based}. One major drawback of these approaches is the use of a stationary camera, which ameliorates the complexity of the problem.

As opposed to traffic surveillance purposes, few studies address the problem from an ADAS perspective, namely estimating speed using monocular images from a camera mounted on the car. Some works estimate the relative speed of other participants in traffic (see, e.g.,~\cite{salahat2017speed} or ~\cite{kampelmuhler2018camera-based}). In the present study we are concerned with estimating the absolute speed of the car the camera is mounted on, also called the ego/forward/longitudinal speed. 

In~\cite{qimin2014amethodology} the authors used sparse optical flow to track feature points on images from a downward-looking camera mounted on the rear axle of the car and achieved a mean error relative to GPS measurement of 0.121 m/s. However, the method works only in restricted conditions and was evaluated on self-collected data at low speed values.
Han~\cite{han2016car} used projective geometry concepts to estimate relative and absolute speed in different case studies. Using black box footages, a maximum of 3\% difference was reported for higher ego-speed values when compared to GPS measurements. The major limitation of this study is the assumption of known distances between stationary objects such as lane markings.
Banerjee et al.~\cite{banerjee2017velocity} used a rather complicated neural network architecture trained on self-collected data and reported an RMSE of 10 mph on the KITTI benchmark~\cite{Geiger2013IJRR}.

In this work we investigate a simple intuitive approach for ego-speed estimation using deep neural network-based optical flow and monocular depth prediction methods. We achieve a RMSE of less than 1 m/s on recordings from the KITTI dataset.

\section{Methods}\label{sect:methods}
\subsection{The KITTI dataset}
The KITTI Vision Benchmark Suite\footnote{\url{http://www.cvlibs.net/datasets/kitti}}~\cite{Geiger2012CVPR, Geiger2013IJRR} is a real-world dataset consisting of 6 hours of traffic scenario recordings captured while driving in and around a mid-size city. The traffic situations range from highways over rural areas to innercity scenes with many static and dynamic objects.

The recording platform consisted of high resolution stereo camera systems (two color and two grayscale video cameras) mounted on top of the vehicle, Velodyne laser scanner that produces more than one million 3D points per second, a high-precision localization system, and a powerful computer running a real-time database housed in the trunk of the car. The cameras, laser scanner and localization system were calibrated and synchronized, providing accurate timestamped ground truth data. The dataset consists of both the raw and rectified, color and grayscale image sequences, laser scans, object labels in the form of 3D tracklets, and online benchmarks for stereo matching and optical flow estimation, 3D visual odometry, and 3D object detection.

The introductory paper~\cite{Geiger2012CVPR} focuses on presenting the benchmark challenges, their creation and use for evaluating state-of-the-art computer vision methods, while the follow-up work~\cite{Geiger2013IJRR} provides technical details on the raw data itself, describing the recording platform, the data format and the utilities.

The KITTI raw dataset is further augmented by additional data and benchmarks presented in subsequent papers. The stereo 2015 / flow 2015 / scene flow 2015 benchmark is described in~\cite{Menze2015CVPR}. This is a novel and realistic dataset with semi-dense scene flow ground truth obtained by annotating 400 highly dynamic scenes (200 training and 200 test) from the KITTI raw data collection using detailed 3D models for all vehicles in motion. Reference~\cite{Uhrig2017THREEDV} presents a dataset derived from the KITTI raw dataset, for training and evaluating depth completion and depth prediction techniques, comprising over 94k images annotated with high-quality semi-dense depth ground truth.

\subsection{Optical flow estimation: {FlowNet2} and \mbox{PWC-Net}}

FlowNet2\footnote{\url{https://github.com/lmb-freiburg/flownet2}}~\cite{ilg2017flownet2} poses the problem of optical flow estimation as an end-to-end supervised learning task and uses convolutional neural networks (CNNs). As a conslidation of the original FlowNet idea, it inherits the advantages of mastering large displacements, correct estimations of fine details, the potential to learn priors for specific scenarios and fast runtimes. FlowNet2, compared to its initial versions, shows large improvements in quality and speed due to focusing on the schedule of presenting the training data during learning, using a stacked architecture including warping of the second image with intermediate optical flow results, and elaborating on small displacements by introducing a subnetwork specialized on small subpixel motions.

PWC-Net\footnote{\url{https://github.com/sniklaus/pytorch-pwc}}~\cite{sun2018pwcnet} makes significant improvements in model size and accuracy over existing CNN models for optical flow. The method was designed according to simple and well-established principles: using learnable feature pyramids, using the warping operation as a layer to estimate large motions and the use of a cost volume layer. PWC-Net is 17 times smaller in size and easier to train than FlowNet2, and achieves state-of-the results on the KITTI 2015 benchmark~\cite{Menze2015CVPR}.

While FlowNet2 achieves impressive performance by stacking basic models into a large capacity model, the much smaller PWC-Net obtains similar or better results by embedding classical principles into the network itself. For more details about the two methods and other optical flow estimation approaches please see the cited works and the references therein.

\subsection{Monocular depth estimation: {MonoDepth} and {MegaDepth}}

MonoDepth\footnote{\url{https://github.com/mrharicot/monodepth}}~\cite{godard2017unsupervised} innovates beyond existing learning based single image depth estimation methods by replacing the use of large quantities and difficult to obtain quality ground truth training data with easier to obtain binocular stereo footage. The authors propose a CNN network architecture that performs unsupervised depth prediction by posing the task as an image reconstruction problem and using a novel training objective that enforces left-right consistency between the disparities of the left and right color images from a calibrated stereo pair. To obtain better accuracy, at test time MonoDepth requires a single input image and computes disparity for its horizontally flipped counterpart too. MonoDepth outperforms supervised methods on the KITTI 2015 benchmark~\cite{Menze2015CVPR}.

MegaDepth\footnote{\url{https://github.com/lixx2938/MegaDepth}}~\cite{li2018megadepth} refers to a large depth dataset generated via modern structure-from-motion and multi-view stereo methods from multi-view Internet photo collections. The authors propose data cleaning methods and augmentation of the data with ordinal depth relations using semantic segmentation, and a corresponding loss function for training. The models trained on MegaDepth exhibit high accuracy and strong generalization ability to novel scenes, including the KITTI dataset. 

One important note to make is that MonoDepth predicts disparity values, which can be converted to depth by multiplying its inverse with the given focal length and baseline between the stereo cameras. On the other hand, the MegaDepth models do not predict metric depth, but ordinal depth defined up to a scale factor. For more details about the two methods and related single-view depth estimation approaches please see the cited works and the references therein.

\subsection{Speed estimation pipeline}
To estimate ego-speed from monocular images of a camera mounted on the car our approach employs a multistage process. In a first step we run one optical flow and one depth estimation algorithm on a given image frame. Then, the results at valid pixels are considered from a predefined crop of the original image, and the mean of the magnitude of optical flow vectors (denoted by OF in the following) is divided by the mean of the disparity (DISP) values. The valid pixels are obtained by imposing thresholds: OF \textgreater{} 0.2 and DISP \textgreater{} 0.01. The next step is the concatenation of the scaled speed estimates over the temporal dimension, i.e. over frames of a video. The aggregated vectors are temporally smoothed using a 1D convolution of size 25 with equal weights. Finally, the resulting smoothed lists are taken for multiple video recordings and a scaling factor is approximated that minimizes the ratio between the ground truth and predicted speed. This scaling factor is used to convert the estimated speed from the image domain to real-world units.

To summarize, the speed estimation pipeline steps are as follows:
\begin{enumerate}
    \item Run optical flow and depth estimation method on image frame.
    \item Compute the scaled speed for the given frame: consider the OF and DISP values at valid pixels from a predefined image crop, and compute the quotient between their means.
    \item Repeat the previous steps for all the frames from a video and apply temporal smoothing.
    \item Repeat the previous step for multiple videos and estimate scaling factor.
\end{enumerate}
Several modifications can be made to the base speed estimation algorithm defined above; these will be experimented in Section~\ref{sect:results}.
\begin{enumerate}[label=(\subscript{E}{{\arabic*}})]
    \item As speed is expected to highly correlate with optical flow, depth information could be disregarded, considering only OF.
    \item Instead of OF, one could consider using the magnitude of horizontal optical flow only (which is expected to highly correlate with the moving speed especially towards the edges of the image).
    \item Temporal smoothing can also be applied at the pixel-level separately for OF and DISP.
    \item As opposed to providing the full frame as input to the neural networks and taking the results on image crops, one could run the methods only on the crops.
    \item Different image crops can be considered when taking the OF and DISP results, including the full frame too.
\end{enumerate}

The images in the KITTI dataset are of $1242\times375$ resolution. Before obtaining the dense optical flow and depth results, the input images need to be resized to the appropriate resolution. We also resize the results of the deep neural networks back to the original resolution as summarized in Table~\ref{table:inputsize}. Our choices provided the best test results when evaluating the methods on KITTI. To run the neural networks we used pre-trained weights provided in the corresponding Github repositories.

\begin{table}[h]
\centering
{\footnotesize
\begin{tabular}{lcccc}
Method & Input resolution & Input resize & Output resize & Model used\textsuperscript{\textasteriskcentered}\\
\hline
FlowNet2  & divisible by 64 & pad with zeros & trim zeros   & FlowNet2\\
PWC-Net   & $1024\times436$ & bilinear interpolation & anti-aliasing & default network\\
MonoDepth & $512\times256$  & anti-aliasing & anti-aliasing & city2kitti\\
MegaDepth & $512\times384$  & anti-aliasing & anti-aliasing & best generalization\\
\hline
\end{tabular}
\caption{{\bf Deep neural network technicalities.} \textsuperscript{\textasteriskcentered}For details check the references and/or the Github repositories.}
\label{table:inputsize}}
\end{table}	

To evaluate our speed estimation pipeline we manually selected 15 recordings of rectified images from the left input color camera of the KITTI dataset~\cite{Geiger2012CVPR}. The list of drive IDs is summarized in Table~\ref{table:kittilist}. These are reresentative videos where the car is moving almost always. We did not include recordings in which the car is stationary from start to finish, as zero speed might highly increase the accuracy of our method.

\begin{table}[h]
\centering
{
\begin{tabular}{cc}
Number & KITTI ID\\
\hline
 1&2011\_09\_26\_drive\_0001\\
 2&2011\_09\_26\_drive\_0002\\
 3&2011\_09\_26\_drive\_0005\\
 4&2011\_09\_26\_drive\_0009\\ 
 5&2011\_09\_26\_drive\_0014\\ 
 6&2011\_09\_26\_drive\_0019\\
 7&2011\_09\_26\_drive\_0027\\
 8&2011\_09\_26\_drive\_0048\\
 9&2011\_09\_26\_drive\_0056\\
10&2011\_09\_26\_drive\_0059\\
11&2011\_09\_26\_drive\_0084\\
12&2011\_09\_26\_drive\_0091\\
13&2011\_09\_26\_drive\_0095\\
14&2011\_09\_26\_drive\_0096\\
15&2011\_09\_26\_drive\_0104
\end{tabular}
\caption{{\bf List of KITTI drive IDs used in our experiments.}}
\label{table:kittilist}}
\end{table}	


In experiment $E_5$ defined previously we evaluate our proposed speed estimation pipeline on three different image crops. They are defined in Table~\ref{table:crops_def} and visualized on two sample images from the KITTI dataset. Reasons for using crops only include the possible unavailability of wide images, or memory and run-time considerations.

\begin{table}[h]
\centering
{
\begin{tabular}{lrrrr}
\multirow{2}{*}{Crop name} &\multicolumn{4}{c}{bounding box}\\
& $x$ & $y$ & $w$ & $h$ \\
\hline
cropB & 720 & 180 & 200 & 120\\
cropG & 700 & 100 & 400 & 240\\
cropR & 640 &  20 & 580 & 340\\
\end{tabular}
\footnotesize
\caption{{\bf Definition of image crops used in our experiments.} $(x, y)$ defines the upper left corner and $w$, $h$ the width and height of the bounding boxes.}
\label{table:crops_def}}
\end{table}	

\begin{figure}[h]
 \footnotesize
 \centering
 \includegraphics[width=.49\textwidth]{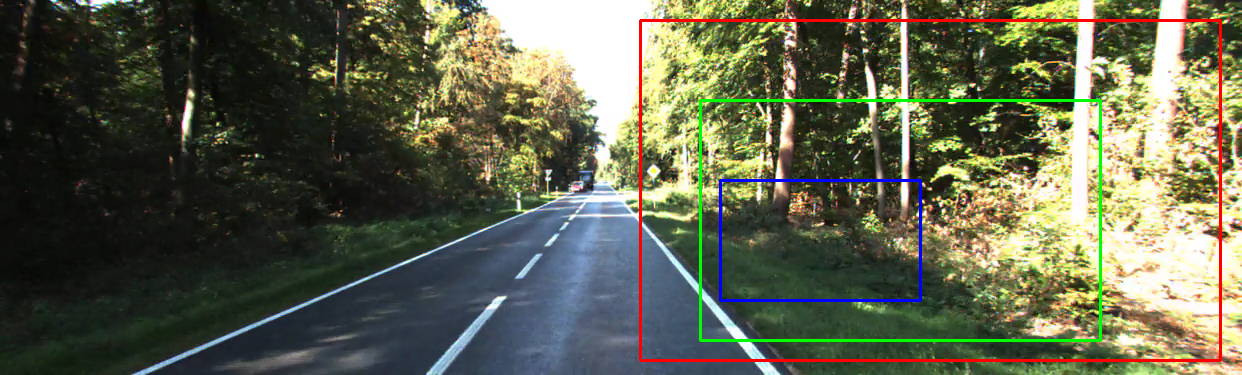}
 \includegraphics[width=.49\textwidth]{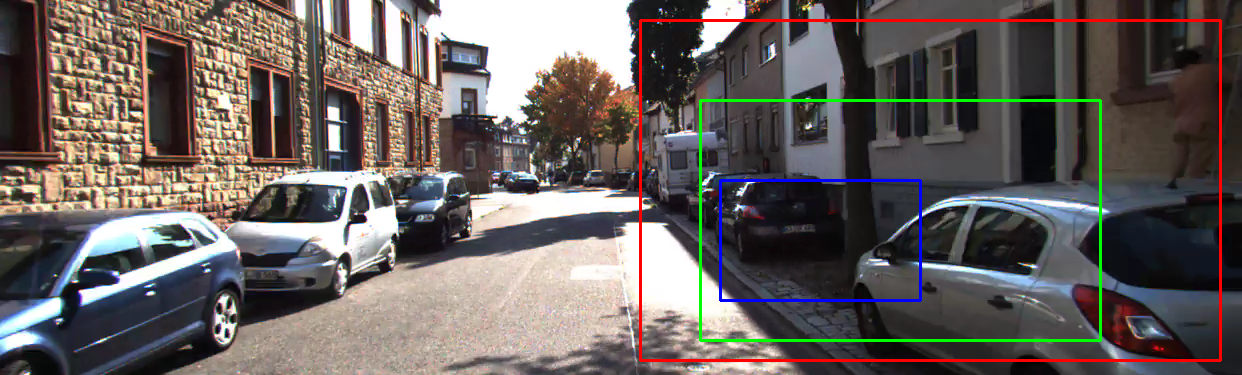}
 \caption{{\bf Image crops used in our speed estimation experiments.} Bounding boxes are overlaid on two sample frames from the KITTI dataset (left: frame 14 of {\it drive\_0027}, right: frame 26 of {\it drive\_0095}). The definition of the bounding boxes is shown in Table~\ref{table:crops_def}.}
 \label{fig:crops}
\end{figure}

\section{Experiments and results}\label{sect:results}
\begin{figure}[h]
 \footnotesize
 \centering
 \includegraphics[width=.49\textwidth]{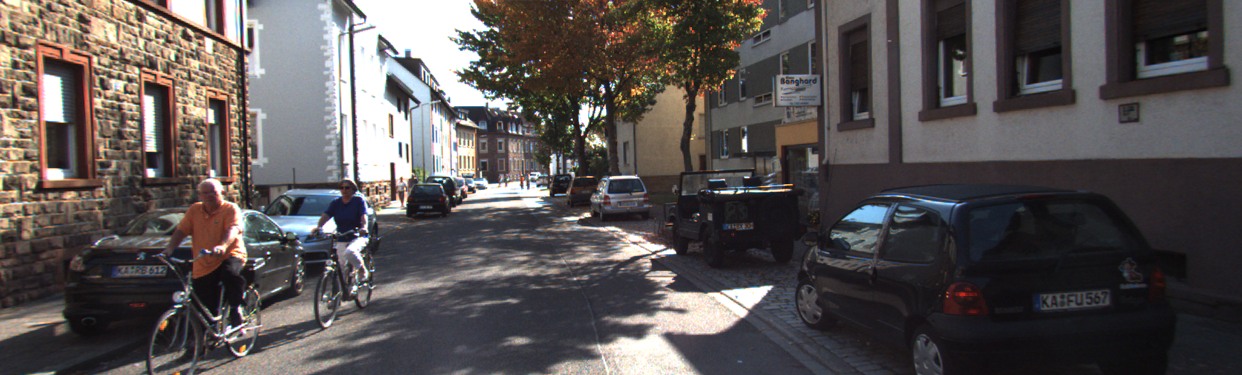}
 
 \includegraphics[width=.49\textwidth]{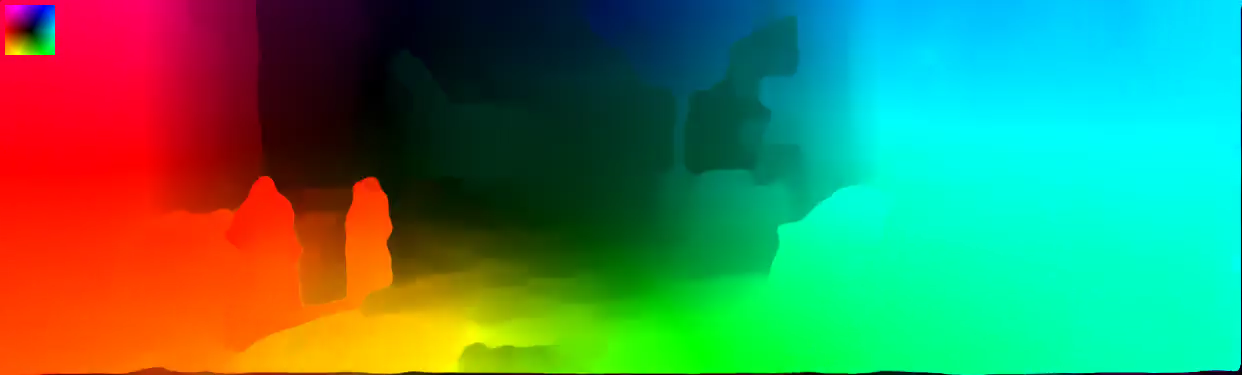}
 \includegraphics[width=.49\textwidth]{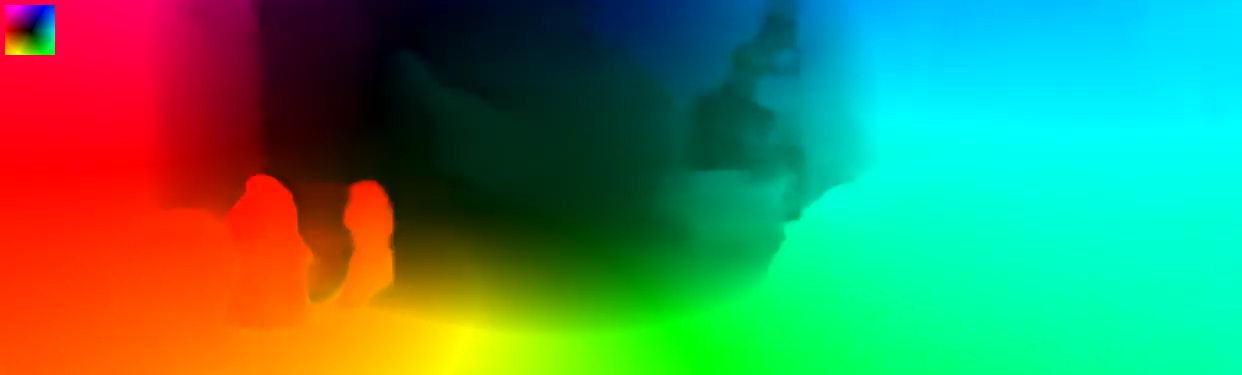}
 
 \includegraphics[width=.49\textwidth]{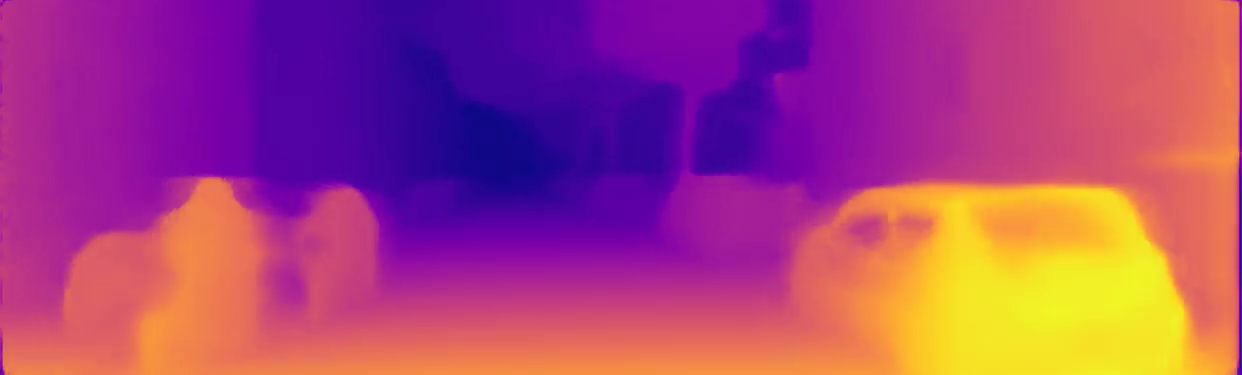}
 \includegraphics[width=.49\textwidth]{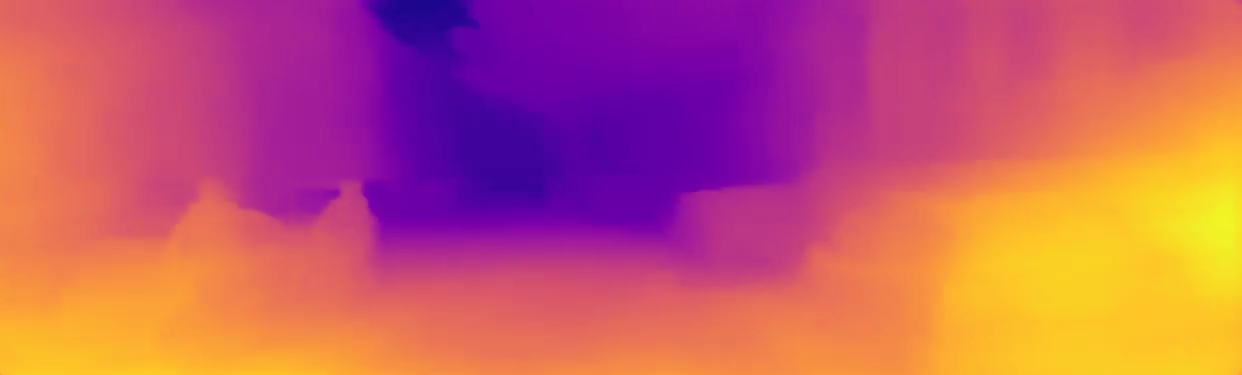}
 \caption{{\bf Sample visualisation of network results.} From left to right and top to bottom: frame 71 of {\it drive\_0095}, FlowNet2, PWC-Net, MonoDepth, MegaDepth. The colored square represents the color coding of optical flow.}
 \label{fig:sample_viz}
\end{figure}

Figure~\ref{fig:sample_viz} shows visualizations of the deep neural network methods used in our experiments on one sample frame from the KITTI dataset. Both optical flow estimation methods produce smooth flow fields with sharp motion boundaries. PWC-Net seems to be more robust against shadow effects. The depth prediction methods also show good visual quality, with MonoDepth able to capture object boundaries and thin structures more reliably.

We also evaluate quantitatively the methods. The optical flow estimation algorithms are evaluated on the 200 training images from the KITTI 2015 benchmark~\cite{Menze2015CVPR}. The results displayed in Table~\ref{table:of_eval_kitti} show that PWC-Net outperforms FlowNet2. The depth prediction algorithms are evaluated on the given 1000 manually selected images from the full validation split of the derived depth prediction and completion KITTI dataset~\cite{Uhrig2017THREEDV}. Table~\ref{table:depth_eval_kitti} shows the results for several error metrics. We can see that MonoDepth achieves better accuracy values compared to MegaDepth. In all four cases the results are in correspondence with those reported in the references presenting the methods.

\begin{table}[h]
\centering
{\renewcommand{\arraystretch}{1.0}
\begin{tabular}{lcc}
Method & AEPE & Fl-all\\
\hline
FlowNet2 & 11.686 & 32.183\%\\
PWC-Net & 2.705 & 9.187\%\\
\end{tabular}
\footnotesize
\caption{{\bf Evaluation of optical flow estimation methods on the KITTI 2015 benchmark~\cite{Menze2015CVPR}.} AEPE: average endpoint error; Fl-all: Ratio of pixels where flow estimate is
wrong by both $\geq$ 3 pixels and $\geq$ 5\%.}
\label{table:of_eval_kitti}}
\end{table}	

\begin{table}[h]
\centering
{\renewcommand{\arraystretch}{1.0}
\begin{tabular}{lcccccc}
Method & RMSE & RMSE(log) & Abs Rel & Sq Rel & log10 & Scale-inv.\\
\hline
MonoDepth & 4.532 & 0.150 & 0.090 & 0.749 & 0.040 & 0.142\\
MegaDepth & 6.719 & 0.336 & 0.322 & 1.994 & 0.124 & 0.289\\
\end{tabular}
\footnotesize
\caption{{\bf Evaluation of depth estimation methods on the manual selection of the  validation split of the derived depth prediction and completion KITTI 2017 dataset~\cite{Uhrig2017THREEDV}.}}
\label{table:depth_eval_kitti}}
\end{table}	

Table~\ref{table:experiments} shows the speed estimation results for experiments $E_1, E_2, E_3, E_4$. We can formulate the following conclusions: in all cases considering depth information as well, as opposed to optical flow alone improves performance ($E_1$), replacing OF with horizontal optical flow results in slightly higher RMSE values ($E_2$), applying temporal smoothing at the pixel level may improve performance but only marginally ($E_3$), applying the deep neural network methods on a smaller image region (crop) -- as opposed to first applying on the original wide frame and then extracting the results from the corresponding crop -- influences performance considerably in a negative way ($E_4$).

\begin{table}[h]
\centering
{\renewcommand{\arraystretch}{1.0}
\begin{tabular}{lcccc}
\multirow{3}{*}{Method} &\multirow{3}{*}{\shortstack[l]{Base\\ pipeline}} & Horizontal & Pixel-level & Methods\\
& & optical flow & smoothing & only on crop\\
&&($E_2$) & ($E_3$) & ($E_4$) \\
\hline
FlowNet2 ($E_1$)     & 2.921 & 3.005 & 2.919 & 3.093 \\
PWC-Net  ($E_1$)     & 2.472 & 2.621 & 2.472 & 3.170 \\
FlowNet2 \& MonoDepth& 2.305 & 2.448 & 2.399 & 2.618 \\
PWC-Net  \& MegaDepth& 1.915 & 2.059 & 1.908 & 3.180 \\
FlowNet2 \& MegaDepth& 2.485 & 2.526 & 2.475 & 2.901 \\
PWC-Net  \& MonoDepth&{\bf1.467}& 1.707 & 1.865 & 2.967 \\
\end{tabular}
\footnotesize
\caption{{\bf Results of speed estimation experiments.} RMSE values are shown using {\it cropG} defined in Table~\ref{table:crops_def}. $E_i, i\in\{1,2,3,4\}$ refers to the experiments defined in the previous section.}
\label{table:experiments}}
\end{table}	

\begin{figure}[!h]
 \footnotesize
 \centering
 \begin{subfigure}{0.7\textwidth}
    \centering
    \includegraphics[width=\textwidth]{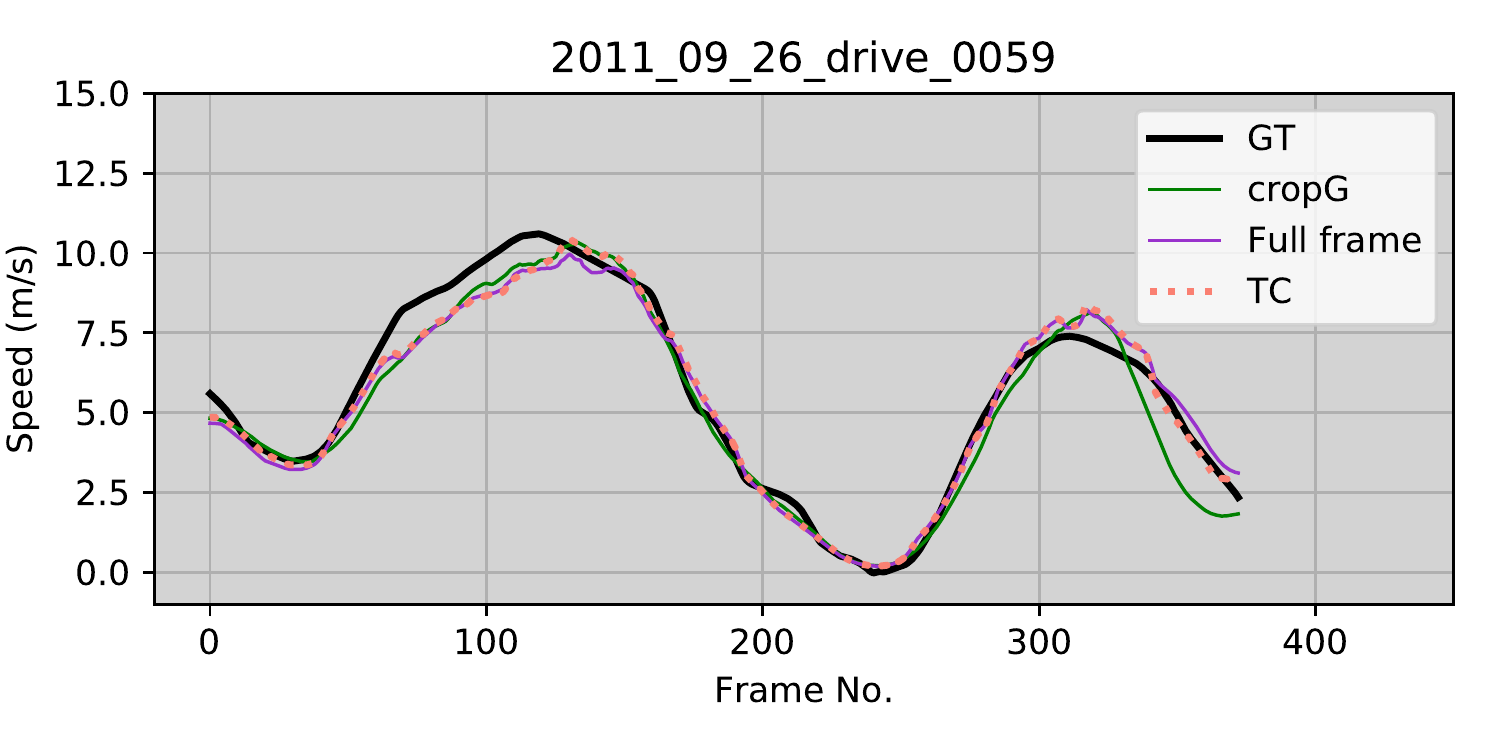}
    \caption{}
    \label{fig:speeda}
  \end{subfigure}
 \begin{subfigure}{0.7\textwidth}
    \centering
    \includegraphics[width=\textwidth]{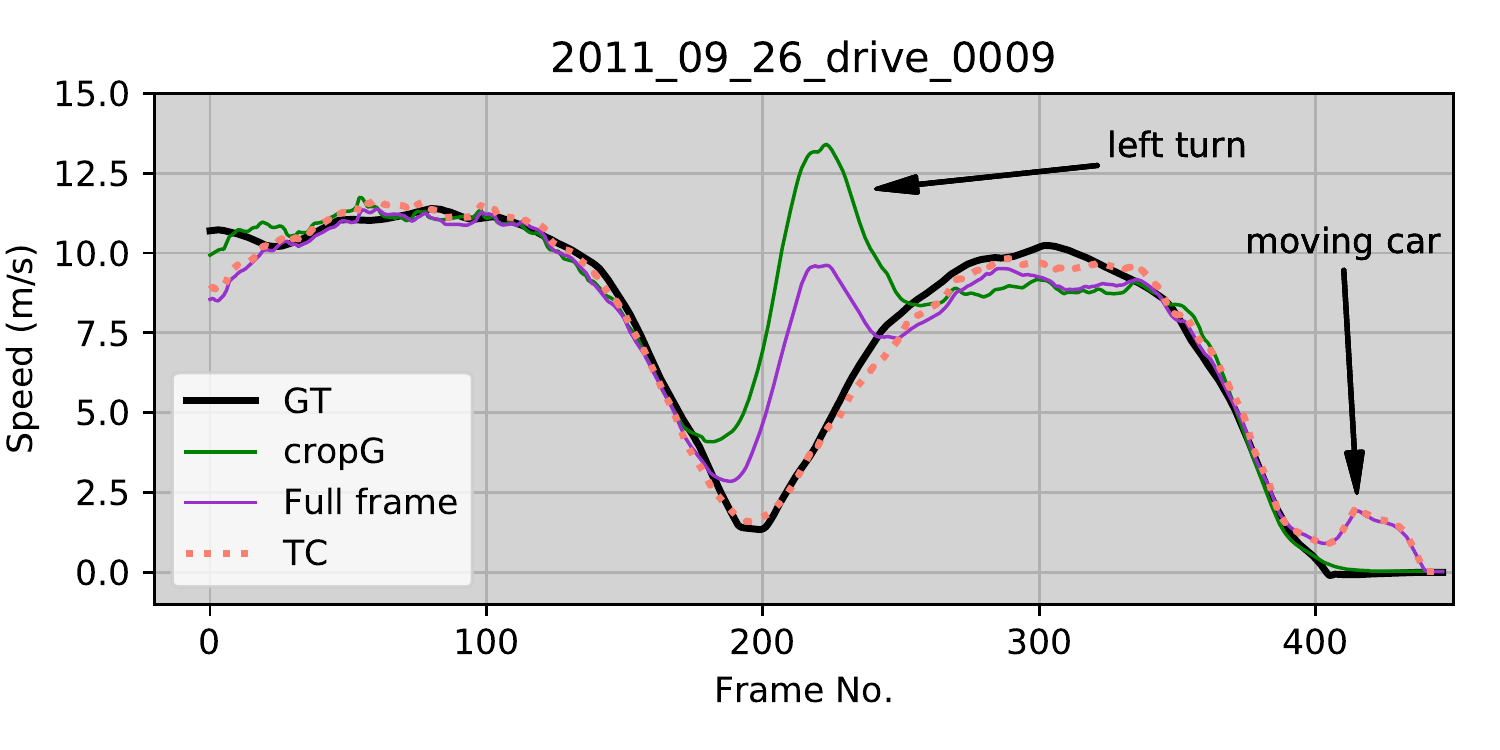}
    \caption{}
    \label{fig:speedb}
  \end{subfigure}
 \caption{{\bf Speed estimation results on two sample KITTI recordings.} The base pipeline was used with the PWC-Net and MonoDepth methods; {\it cropG} is defined in Table~\ref{table:crops_def} and {\it Full frame} refers to the full wide image. TC refers to compensating for car turning events. For more details see text.}
 \label{fig:speed}
\end{figure}

Inspecting Table~\ref{table:experiments} we can see that the best results are obtained when PWC-Net is combined with MonoDepth. Figure~\ref{fig:speed} shows speed estimation results in this case using the base pipeline for two KITTI recordings. Our simple method is able to capture speed changes with low error in straight travel scenarios (Figure~\ref{fig:speeda}), but having difficulty in cases when the car is taking a turn (around frame 200 on Figure~\ref{fig:speedb} speed decreases as the car is turning left, yet optical flow increases on the right side -- and in {\it cropG} too -- of the wide KITTI images).
 
The results of experiment $E_5$ are illustrated in Table~\ref{table:crop_res}. We evaluate the base speed estimation algorithm on the three image crops from Table~\ref{table:crops_def} and using the full frame as well. We can see that speed estimation accuracy improves in general as the size of the image used is increasing. Furthermore, note that similarly to the previous experiments, the best results are obtained by the PWC-Net -- MonoDepth combination. When the full frame is used errors moderately decrease at car turning events (around frame 200 on Figure~\ref{fig:speedb}), but overestimations are still present in dynamic scenes due to the motion of other cars for instance (after frame 400 on Figure~\ref{fig:speedb}).

\begin{table}[h]
\centering
{\renewcommand{\arraystretch}{1.0}
\begin{tabular}{lcccc:c}
Method & {\it cropB} & {\it cropG} & {\it cropR} & Full frame & TC\\
\hline
FlowNet2 \& MonoDepth & 2.170 & 2.305 & 2.558 & 2.370 & 2.138\\
PWC-Net  \& MegaDepth & 2.363 & 1.915 & 2.015 & 1.786 & 1.445\\
FlowNet2 \& MegaDepth & 2.671 & 2.485 & 2.583 & 2.544 & 2.125\\
PWC-Net  \& MonoDepth & 1.735 & 1.467 & 1.505 & {\bf1.178} & {\bf0.977}\\
\end{tabular}
\footnotesize
\caption{{\bf Results using different image crops.} RMSE values are shown for experiment $E_5$. Image crops are defined in Table~\ref{table:crops_def}. TC refers to commpensating for car turning events (for details see text).}
\label{table:crop_res}}
\end{table}	

In order to decrease speed overestimations at car turning events we experimented with a modification of the base algoritmic pipeline. In such cases the average of horizontal optical flow of the left part of the image has the same sign as the average from the right side. Whenever this condition is true, instead of the mean of the optical flow magnitude for the full wide frame, the absolute value of the difference between the means of horizontal optical flow of the left and right sides is computed, and divided by the mean disparity corresponding to the whole frame. Applying this compensation method for turning events (TC) decreases the RMSE to under 1 m/s, as shown in Table~\ref{table:crop_res}. This performance improvement is illustrated by Figure~\ref{fig:speedb} as well.

\section{Discussion and conclusion}\label{sect:discussion_conclusion}
In this work we investigated a simple algorithm for ego-speed estimation using state-of-the-art deep neural network based optical flow estimation and monocular depth prediction on images from a camera mounted on the moving car. We relied on the intuition that optical flow magnitude is highly correlated with the moving speed of the observer and that the closer objects are to the observer the faster they appear to be moving. We defined a base algorithmic pipeline for speed estimation, evaluated several modifications to it and achieved a RMSE of less than 1 m/s on a subset of the widely used KITTI dataset~\cite{Geiger2012CVPR, Geiger2013IJRR}.

We investigated two optical flow estimation (FlowNet2~\cite{ilg2017flownet2} and PWC-Net~\cite{sun2018pwcnet}) and two depth estimation (MonoDepth~\cite{godard2017unsupervised} and MegaDepth~\cite{li2018megadepth}) algorithms. We evaluated these on ground truth data from the KITTI dataset and saw that PWC-Net and MonoDepth achieved better performance in several error metrics. The reason for this is probably some combination of the following: the MonoDepth model used in our work was fine-tuned on data from KITTI, MegaDepth does not predict metric depth but ordinal depth, it seems that during training of the PWC-Net model KITTI data was used as well. Nonetheless, the conclusion is that better performance optical flow and depth estimation methods result in better speed estimates. Besides, continuous efforts are made to improve these two fundamental computer vision algorithms, including their joint training in an unsupervised manner (see, e.g.,~\cite{zou2018dfnet}). Accordingly, fine-tuning to arbitrary images becomes easily accessible without the need for difficult to obtain ground truth labels.

We also evaluated several modifications to our proposed base algorithm and formulated meaningful conclusions. Experiment $E_1$ showed that optical flow alone reached reasonable performance, but using depth information too improved the results. Experiment $E_2$ illustrated that using horizontal optical flow gives lower performance than using the magnitude of the optical flow vectors. In experiment $E_3$ we saw that applying temporal smoothing at the pixel level (as opposed to applying it after image-level aggregation) can only marginally improve performance in some cases. Experiment $E_4$ illustrated that applying the deep neural network methods only on smaller crops of the original wide images significantly influences results in a negative manner. Experiment $E_5$ compared the base speed estimation pipeline using different image crops and proved that using larger regions is better, with the full frame providing the best overall result.

There are two limitations to our method that need to be mentioned. On one hand, speed is erroneously estimated when the car is turning. However, in such cases estimation errors can be corrected by taking into account that horizontal optical flow from the left and right side of the wide images have the same direction (see the last column of Table~\ref{table:crop_res} and Figure~\ref{fig:speedb}). On the other hand, the proposed method is most reliable when the background is static. For example in heavy traffic scenarios when the surrounding cars are moving as well, the correlation of optical flow with ego-speed might be small and speed can be over- (see Figure~\ref{fig:speedb}) or underestimated. In such scenarios combining monocular depth estimation with semantic segmentation~\cite{jiang2018self-supervised} represents one promising direction; and the estimation of relative speed can help, which is another problem where recent advances are being made (see, e.g.,~\cite{salahat2017speed} or~\cite{kampelmuhler2018camera-based}). One might also argue that the deep neural network methods providing the best speed estimates in our case were fine-tuned on ground truth data from KITTI. But, as explained above, efforts are being made to train such methods on unlabelled data \cite{zou2018dfnet}.

To improve our simple method future works can treat the task as a regression problem and adopt for example a lightweight multilayer perceptron using as input the aggregated optical flow and depth results from different smaller regions of the original larger image(s). Another possibility is the exploitation of the more sophisticated convolutional neural network, which, however would possibly require a larger amount of training data. We close this work by noting that due to the recent and ongoing advancements in deep learning, monocular vision-based approaches are a promising direction for ego-speed estimation, and autonomous driving in general.


\section*{Acknowledgements}
The project has been supported by the European Union, co-financed by the European Social Fund (EFOP-3.6.2-16-2017-00013 and EFOP-3.6.3-VEKOP-16-2017-00002). The author acknowledges Andr\'as {L\H orincz}, his PhD supervisor, for his expertise and efforts that promoted the prosperous progression of this project. The author would also like to thank Kinga B. {Farag\'o} for her help and cooperation in supervising the work of two students, and Szilvia Szeier and Kevin A. {Harty\'anyi} for their dedication in preparing their scientific student conference paper.  Finally, the author expresses his gratitude to colleagues and members of the NIPG (Neural Information Processing Group) for their background support.

\bibliographystyle{abbrv}
\bibliography{speed}
\end{document}